\documentclass[letterpaper, 10pt, conference]{ieeeconf}

\IEEEoverridecommandlockouts
\usepackage{graphics}      % For pdf, bitmapped graphics files
\usepackage{epsfig}        % For postscript graphics files
\usepackage{mathptmx}      % Times-like font for math
\usepackage{times}         % Times font
\usepackage{amsmath,amssymb}
\usepackage{xspace}
\usepackage{booktabs}      % \toprule, \midrule, \bottomrule
\usepackage[table]{xcolor}
\usepackage{colortbl}
\usepackage{algorithm}
\usepackage{algpseudocode}
\usepackage{tabularx}
\usepackage{makecell}
\usepackage{tikz}
\usetikzlibrary{positioning,arrows.meta,calc}
\usepackage{subcaption}    % Modern sub-figures
\definecolor{citecolor}{HTML}{0071BC}
\definecolor{linkcolor}{HTML}{ED1C24}
\definecolor{urlcolor}{HTML}{D02090}
\definecolor{brandeisblue}{rgb}{0.0, 0.44, 1.0}

\definecolor{LightCyan}{rgb}{0.88,1,1}
\definecolor{LightRed}{rgb}{1,0.5,0.5}
\definecolor{LightYellow}{rgb}{1,1,0.88}
\definecolor{Grey}{rgb}{0.75,0.75,0.75}
\definecolor{DarkGrey}{rgb}{0.55,0.55,0.55}
\definecolor{DarkGreen}{rgb}{0,0.65,0}
\definecolor{baselinecolor}{gray}{.9}
\definecolor{shadecolor}{RGB}{150,150,150}

\colorlet{darkgreen}{green!65!black}
\colorlet{darkblue}{blue!75!black}
\colorlet{darkred}{red!80!black}
\definecolor{lightblue}{HTML}{0071bc}
\definecolor{lightgreen}{HTML}{39b54a}
\definecolor{deemph}{gray}{0.6}

\usepackage[
    pagebackref=false,
    breaklinks=true,
    letterpaper=true,
    colorlinks=true,
    bookmarks=false,
    citecolor=citecolor,
    linkcolor=linkcolor,
    urlcolor=brandeisblue
]{hyperref}

\usepackage{array}        % Extended column specs (e.g., >{\centering\arraybackslash}p{.})
\usepackage{multirow}     % Multi-row cells
\usepackage{siunitx}      % Numeric alignment with S columns
\renewcommand{\paragraph}[1]{\vspace{0.0em}\noindent\textbf{#1}}
\newcommand{\method}{{Flex}\xspace}

\title{\LARGE \bf
Towards Efficient and Effective Multi-Camera Encoding for End-to-End Driving
}

\author{
Jiawei Yang$^{\pi,*}$, 
Ziyu Chen$^{\rho,*}$, 
Yurong You$^{*}$, 
Yan Wang$^{*}$, 
Yiming Li$^{*}$, 
Yuxiao Chen$^{*}$,
Boyi Li$^{*}$, \\
Boris Ivanovic$^{*}$, 
Marco Pavone$^{\rho,*}$, 
Yue Wang$^{\pi,*}$ \\[0.3em]
$^{\pi}$USC Physical Superintelligence (PSI) Lab \quad
$^{\rho}$Stanford University \quad
$^{*}$NVIDIA Research \\[0.3em]
Project Page: \href{https://jiawei-yang.github.io/Flex}{Flex}
}

\begin{document}
\maketitle
\thispagestyle{empty}
\pagestyle{empty}

%%%%%%%%%%%%%%%%%%%%%%%%%%%%%%%%%%%%%%%%%%%%%%%%%%%%%%%%%%%%%%%%%%%%%%%%%%%%%%%%
\begin{abstract}

We present \method, an efficient and effective scene encoder that addresses the computational bottleneck of processing high-volume multi-camera data in end-to-end autonomous driving. \method employs a small set of learnable scene tokens to jointly encode information from all image tokens across different cameras and timesteps. By design, our approach is geometry-agnostic, learning a compact scene representation directly from data without relying on the explicit 3D inductive biases, such as Bird-Eye-View (BEV), occupancy or tri-plane representations, which are common in prior work. This holistic encoding strategy aggressively compresses the visual input for the downstream Large Language Model (LLM) based policy model. Evaluated on a large-scale proprietary dataset of 20,000 driving hours, our \method achieves 2.2$\times$ greater inference throughput while improving driving performance by a large margin compared to state-of-the-art methods. Furthermore, we show that these compact scene tokens develop an emergent capability for scene decomposition without any explicit supervision. Our findings challenge the prevailing assumption that 3D priors are necessary, demonstrating that a data-driven, joint encoding strategy offers a more scalable, efficient and effective path for future autonomous driving systems.

\end{abstract}

%%%%%%%%%%%%%%%%%%%%%%%%%%%%%%%%%%%%%%%%%%%%%%%%%%%%%%%%%%%%%%%%%%%%%%%%%%%%%%%%
\section{INTRODUCTION}

Vision-Language-Action (VLA) models~\cite{brohan2022rt,kim2024openvla,gao2024survey,tian2025drivevlm}, which use Large Language Models (LLMs) as powerful policy heads, have been revolutionizing end-to-end autonomous driving. A key challenge in this new paradigm is to encode high-bandwidth visual data~\cite{ivanovic2025efficient}, \textit{efficiently} and \textit{effectively}. A self-driving vehicle's multi-camera rig usually collects tens of images per second, resulting in thousands or tens of thousands of visual tokens per forward. Processing this massive data stream imposes a significant computational load on LLM-based policy models, leading to a critical bottleneck.

\begin{figure}
    \centering
    \includegraphics[width=1\linewidth]{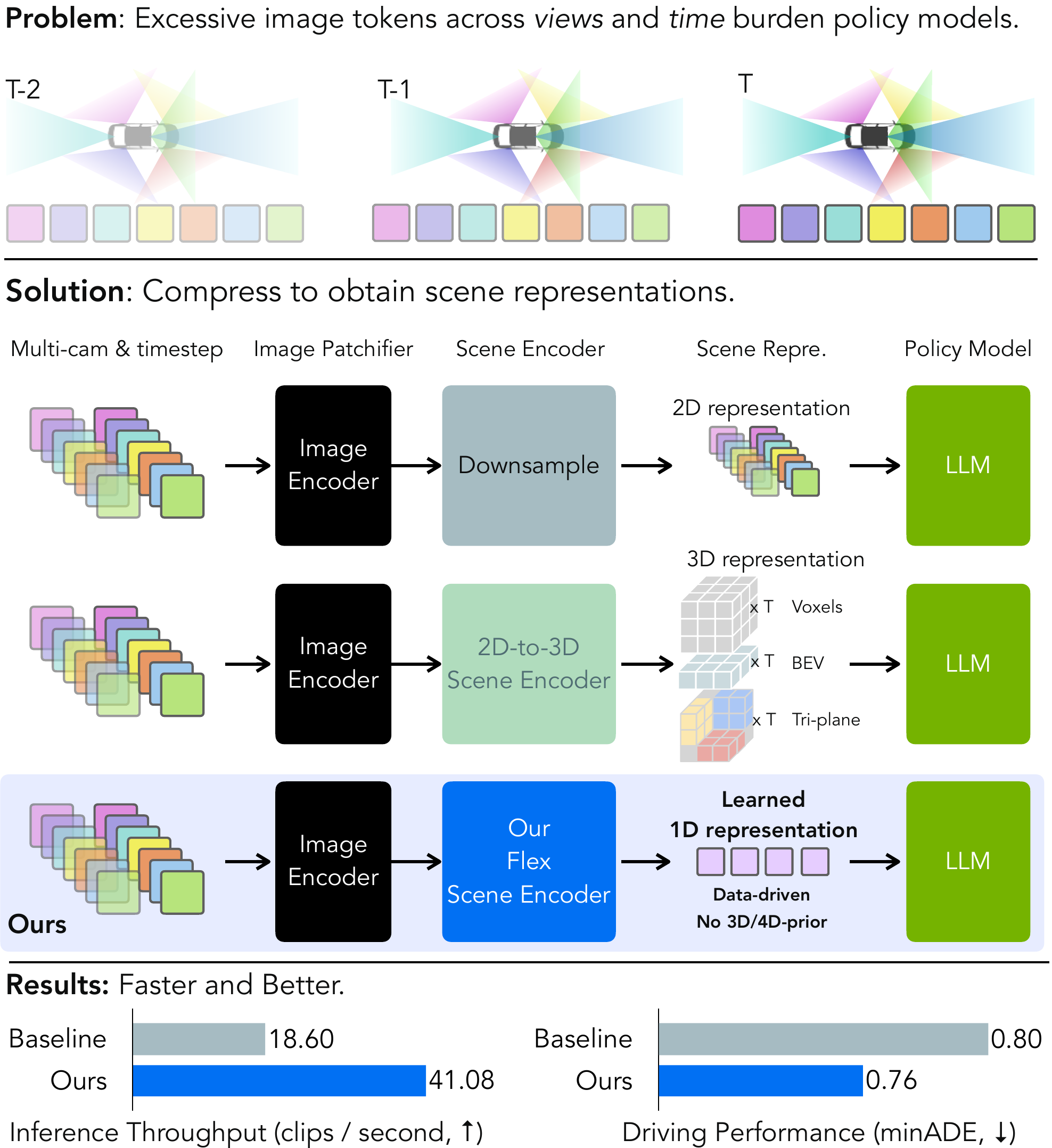}
    \caption{\textbf{Problem, solution, and results.} Multi-camera inputs across time generate excessive tokens that overload policy models. Prior methods compress tokens into 2D or 3D scene representations with limited efficiency and fixed granularity. Our \method scene encoder learns compact scene representations directly from data, without 3D/4D priors or camera poses. It achieves both higher efficiency (41.08 vs. 18.60 clips/s) and better driving accuracy (0.76 vs. 0.80 minADE$_6$).}
    \label{fig:fig1}
\end{figure}

Redundancy lies at the core of this problem. In a typical driving scenario, significant \textit{spatial} overlap exists between adjacent wide field-of-view cameras, while \textit{temporal} continuity introduces further repetition between consecutive frames (see Fig.~\ref{fig:fig1} top). Naive per-image encoding increases latency and memory and, more importantly, fails to exploit the spatiotemporal \textit{redundancy}. Designing a scene encoder that can leverage and compress this redundancy is therefore essential for both efficiency and effectiveness.

Prior efforts often manage this complexity by imposing strong 3D inductive biases.  Methods based on bird-eye-view (BEV) planes~\cite{li2024bevformer,hu2023planning}, voxels~\cite{tong2023scene}, triplanes~\cite{huang2023tri,ivanovic2025efficient}, or hexplanes~\cite{cao2023hexplane} explicitly structure the visual information into a quasi-3D representation, presuming these geometric priors are essential for coherent scene understanding. While these approaches can improve efficiency by factorizing the scene, they also impose a rigid structure that may cap the model's performance. The development of deep learning has repeatedly shown that replacing inductive biases with scalable, data-driven solutions often leads to better performance and generalization~\cite{darcet2023vision}. This trend motivates us to explore a simpler, more scalable alternative, shown in Fig.~\ref{fig:fig1}.

In this paper, we present \method, a \textit{flexible} and geometry-agnostic Transformer scene encoder that jointly encodes images from all views and timesteps into a compact set of scene tokens. Concretely, \method initializes a small set of learnable scene tokens, concatenates them with all image tokens across cameras and time, and runs a lightweight Transformer~\cite{vaswani2017attention} encoder that jointly processes all tokens using self-attention layers. After encoding, we discard the image tokens and keep only the updated scene tokens as the representation passed to the LLM-based policy model. This creates an information-seeking bottleneck that forces cross-view and cross-time compression, allowing the model to learn the optimal representation directly from data, avoiding any explicit 3D priors.

Our design is intentionally made simple, targeting both \textit{efficiency} and \textit{effectiveness}. Efficiency comes from reducing the token budget passed to the policy model by about \textit{3$\times$} to \textit{20$\times$}, which is typically the dominant training and inference cost. Effectiveness comes from the \textit{joint} encoding: the scene tokens attend to all images across views and timesteps, so redundancy suppression occurs at the scene level rather than image level. In contrast, per-image queries, like those used in Q-former~\cite{li2023blip}, cannot enforce global coordination and tend to preserve global redundancy.

Our approach is surprisingly effective. We evaluate \method on a large-scale proprietary dataset of \textbf{20,000} driving hours. Compared to state-of-the-art image encoding strategies, our \method achieves \textbf{2.2$\times$} the inference throughput of baseline methods while improving driving performance by a large margin. These results suggest that a simple, data-driven scene encoder can be more effective. Importantly, we find that jointly encoding the entire scene with self-attention is critical to performance, and surprisingly, our compressed token representation shows an emerging ability to focus on destination, lane markers and safety-critical areas. We hope our work will provide a new perspective on scene encoding for end-to-end driving.

\section{Related work}

\subsection{End-to-end driving and camera visual encoding}

End-to-end autonomous driving has evolved from modular stacks to architectures that fuse perception and planning into a single differentiable system. Recent planning-oriented designs unify tasks with shared queries and BEV features (e.g., UniAD~\cite{hu2023planning}), or learn spatial–temporal features for perception–prediction–planning in one network (e.g., ST-P3~\cite{hu2022st}), yet they still commit to explicit 3D structures, thus requiring precise camera pose information. Vectorized end-to-end approaches (e.g., VADv2~\cite{chen2024vadv2}) consume multi-view image sequences directly and output actions, but do not address aggressive token-level compression across all cameras and timesteps. Our work complements these systems: we focus on the \emph{camera encoding stage}, targeting a compact set of scene tokens learned jointly across views and time, without imposing a particular geometric scaffold. 

\subsection{Scene representations for autonomous driving}

The massive volume of camera inputs creates a computation bottleneck, leading recent works to adopt geometry-grounded representations for compression.
Perception models convert images to Bird’s-Eye-View (BEV) or volumetric occupancy as an intermediate representation. For example, PETR~\cite{liu2022petr} and PETRv2~\cite{liu2023petrv2} inject 3D position encodings for camera-only 3D detection. BEVFormer~\cite{li2024bevformer} uses grid-shaped BEV queries with spatiotemporal Transformers to aggregate multi-view, multi-timestep perceptions. BEVFusion~\cite{liang2022bevfusion} unifies multi-modal features in BEV efficiently; SurroundOcc~\cite{wei2023surroundocc} predicts 3D occupancy from cameras via 2D-to-3D attention. 

Beyond BEV and occupancy, neural field methods propose compact factorized 3D/4D representations.  
Tri-planes factorize features into three axis-aligned planes for efficient 3D representation~\cite{ivanovic2025efficient,huang2023tri}, which are further extended to dynamic scenes by HexPlane and K-Planes ~\cite{chan2022efficient,cao2023hexplane,fridovich2023k}. Instant-NGP~\cite{muller2022instant} and its variants use multi-resolution hash encodings to accelerate field learning \cite{yang2023emernerf}.
These representations improve speed and structure, but they predefine the basis and routing of information. Explicit 3D/4D representations also demand accurate camera poses and synchronized sensors, raising data requirements. Moreover, the pre-set granularity (e.g., grid or voxel size) constrains the information density, and camera perspective further skews this density: near objects occupy many pixels, whereas distant objects collapse into very few. Thus, fixed grid/voxel designs cannot adapt to the highly non-uniform information density in perspective images. In contrast, we minimize geometry-specific priors and compress via data-driven joint attention over the full image set. Our encoder uses a small set of learned scene queries to discover structure directly from data, without committing to any fixed 3D decomposition, and naturally adapts to the non-uniform density of perspective images.  

\subsection{Learned representations} 

A separate thread in computer vision learns a fixed-size query set to summarize large inputs. Perceiver~\cite{jaegle2021perceiver} and Perceiver IO~\cite{jaegle2021perceiverio} introduce asymmetric cross-attention from a small latent array to high-dimensional inputs, decoupling compute from input length. Flamingo uses a Perceiver Resampler to map image/video features to a small, learned set of visual tokens for VLMs~\cite{alayrac2022flamingo}. TokenLearner adaptively extracts few informative tokens from images~\cite{ryoo2021tokenlearner}. The encoder-decoder architecture in TiTok~\cite{yu2024image} and LVSM~\cite{jin2024lvsm} leverage learnable queries to enable compact novel view synthesis and auto-encoding tasks. STORM~\cite{yang2024storm} use learnable tokens to capture the underlying motion biases of the scene. Our scene tokens follow this latent-query paradigm but differ from them in two ways critical for driving: (i) we \emph{jointly} attend over \emph{all} cameras and timesteps at once with self-attention (not per-image or cross-attention-based), and (ii) we show that planning-centric compression naturally produces tokens that are not only compact but also action-relevant. This design yields stronger efficiency–accuracy trade-offs and facilitates end-to-end training.

\begin{figure*}
    \centering
    \includegraphics[width=1\linewidth]{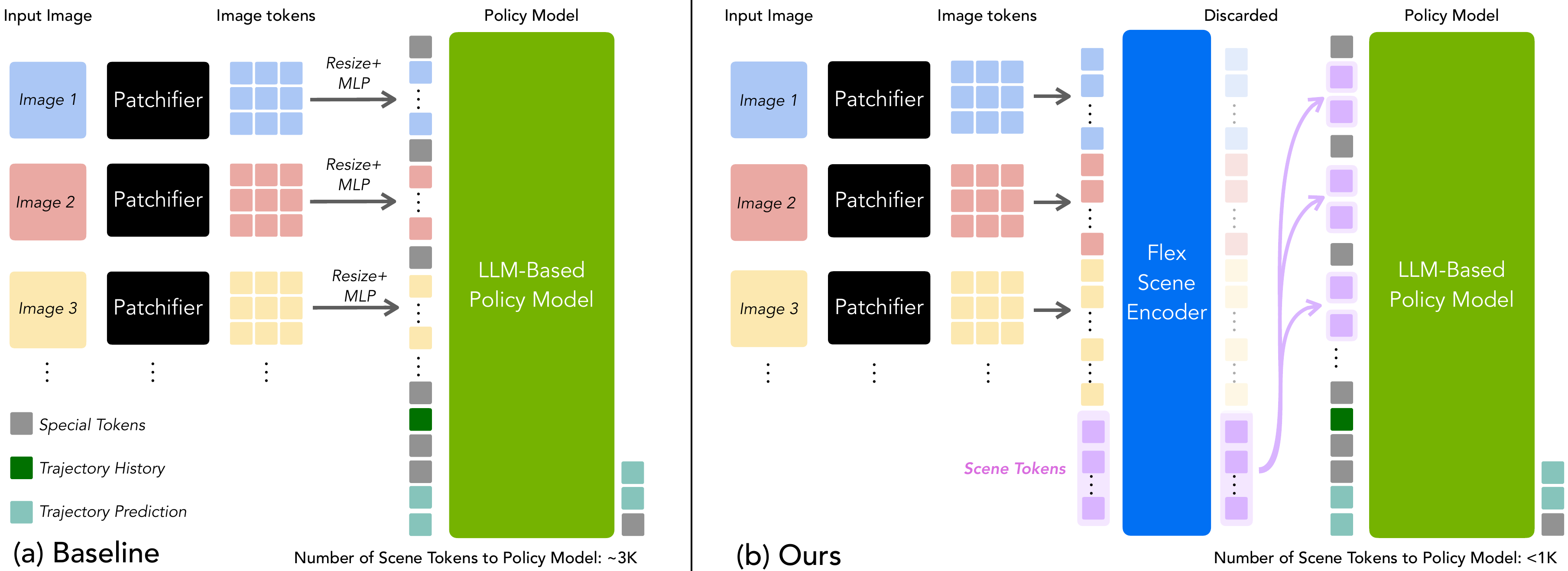}
    \vspace{-1.2em}
    \caption{\textbf{Method overview.}
    In (a) baseline (left), thousands of image tokens from all cameras and timesteps are generated, overloading the policy model. In (b) \method (right), a \method Scene Encoder compresses image tokens into a compact set of scene tokens as scene representations. Only this small set of tokens is used as the scene representations. Special tokens encode the start and end of each modality,  camera type and timestamp information.}
    \label{fig:method}
    \vspace{-1.5em}
\end{figure*}

\section{Method}

Our work introduces \method, a scene encoder designed for Vision-Language-Action (VLA) models in autonomous driving. The overall objective is to learn a policy $\pi$ that predicts a future ego-vehicle trajectory $\mathbf{Y}$ given a sequence of  observations $\mathbf{O}$.
Our primary contribution is a new scene encoder that produces a compact, yet highly effective, scene representation from the massive amount of visual inputs, which improves efficiency during training and inference, and enhances driving performance of the policy model.

\subsection{Vision-Language-Action Model}
\label{sec:va_model}
We first establish a VLA architecture as our baseline. This model processes a sequence of RGB images and past trajectory data to auto-regressively predict a future trajectory. Fig~\ref{fig:method}-(a) shows the overview.

\paragraph{Setup and Notation.} Let $\mathcal{C}=\{1,\dots,C\}$ be camera indices and $\mathcal{T}=\{1,\dots,T\}$ be past timesteps within a fixed window (covering around 2s). The observations $\mathbf{O}$ consist of multi-camera (C), multi-timestep (T) images $\{I_{c,t}\in\mathbb{R}^{H\times W\times 3}\mid c\in\mathcal{C},\,t\in\mathcal{T}\}$ and the most recent ego-state history $\mathbf{s}_{T-h:T}$ (e.g., pose/velocity), where $h$ is the time horizon. The target is a future trajectory $\mathbf{Y}=\{\mathbf{y}_{t}\}_{t=T+1}^{T+h}$ in ego coordinates, where $h$ is the prediction horizon.

\paragraph{Scene Representations.}
For the baseline model, scene representations are derived directly from image tokens. Each input image $I_{c,t} \in \mathbb{R}^{H \times W \times 3}$ is first divided into patches and processed by an image encoder (e.g., DINOv2~\cite{oquab2023dinov2}), which we term ``patchifier'' following VGGT~\cite{wang2025vggt}, to produce a set of $N$ tokens, typically $N=640$ in our setting. These tokens are then bilinearly downsampled to a more manageable number, such as 160 tokens per image. After being resized, they are passed through a 2-layer MLP projection layer $\phi_{proj}$ to match the hidden dimension $D_{llm}$ of the LLM policy head. The final scene representation for the baseline, $\mathbf{S}_{\text{baseline}}$, is a direct concatenation of all processed image tokens from all $C$ camera views and $T$ timesteps:
\begin{equation}
\mathbf{S}_{\text{baseline}} = \text{Concat}\left(\{\phi_{proj}(\text{Resize}({\text{patchifier}}(I_{c,t})))\}_{c,t}\right)
\end{equation}
This results in a long sequence of $C \times T \times 160$ tokens, which imposes a significant computational load on the policy head.

\paragraph{Trajectory Encoding.}
We adopt two complementary tokenization methods.  
(i) \textbf{Continuous history.} The ego-vehicle’s recent trajectory $\mathbf{s}_{T-h:T}$ is encoded by an MLP $\phi_{hist}$ into a single embedding $\mathbf{h}_{hist}$, compressing history into one token.  
(ii) \textbf{Discrete future.} The future trajectory $\mathbf{Y}$ is discretized into a vocabulary of waypoint tokens, added to the LLM embeddings. The policy model then predicts the future as an auto-regressive next-token generation task.  
This encoding proved most effective in our early experiments.

\paragraph{Policy Head.}
We initialize our policy model from a pretrained LLM backbone. This allows us to leverage the rich world knowledge and reasoning capabilities acquired during its pretraining on internet-scale data. The policy model is an auto-regressive Transformer that takes the scene representation and history tokens as context to predict the sequence of future. The model is trained with a standard cross-entropy loss on the predicted future trajectory tokens. Given the large number of scene tokens, this policy model is the dominant component in both training and inference costs. 

\subsection{\method{} Scene Encoder}
\paragraph{Overview.}
Our \method{} encoder is designed to reduce the excessive number of image tokens in the scene representation. Our core insight is that the concatenated image tokens, $\mathbf{S}_{\text{baseline}}$, are highly redundant due to significant spatial overlap between camera views and temporal continuity between frames. To address this, \method{} \emph{jointly} encode all views and timesteps into a \emph{small} set of scene tokens, forcing the model to leverage cross-spacetime correspondences. This design promotes both efficiency and the learning of a more effective scene representation. Fig~\ref{fig:method}-(b) shows the overview.

\paragraph{Joint Scene Encoder.}
Our scene encoder is a lightweight Transformer that operates on all image tokens holistically. As in the baseline, input images are first patchified by a DINOv2 encoder~\cite{oquab2023dinov2}. We then add two sets of positional embeddings to these image tokens: a timestep embedding $PE_t^{\text{time}}$ and a camera embedding $PE_c^{\text{cam}}$. The timestep embeddings are similar to those in DiT~\cite{peebles2023scalable}, while the camera embeddings are learnable vectors indexed by camera types.

We initialize a set of $K$ learnable scene tokens $\mathbf{S}^{(0)}\in\mathbb{R}^{K\times d}$, which act as queries. These scene tokens are prepended to the sequence of all image tokens from all cameras and timesteps. A $L$-layer Transformer encoder $E_\theta$ (we use $L=8$) computes \textit{full self-attention} updates over this combined sequence:
\begin{equation}
\Big[\mathbf{S}^{(L)};\,\mathbf{X}^{(L)}\Big]
\;=\;
E_\theta\!\left(\big[\mathbf{S}^{(0)};\,\mathbf{X}\big]\right),
\qquad
\mathbf{S}^{(L)} \in \mathbb{R}^{K\times d}.
\end{equation}
After the final layer, we \emph{discard} $\mathbf{X}^{(L)}$ and use only the updated scene tokens $\mathbf{S}^{(L)}$ as the scene representation, further denoted as $\mathbf{S}_{\text{\method{}}}$. After a linear projection to $D_{\text{llm}}$, these $K$ tokens feed the policy model.

\begin{table*}[h!]
\centering
\caption{\textbf{System-level comparison}. We compare $\mathrm{minADE}_6$ (lower is better), policy input length (\#Scene Tokens), training cost (A100 GPU hours), and inference throughput (clips/s). \method achieves the best effectiveness–efficiency trade-off, with lower error, fewer tokens, faster training, and higher throughput. Patch sizes are denoted as ($p_x-p_y-p_z$) for triplanes~\cite{ivanovic2025efficient}.}
\label{tab:main}

\vspace{-4pt}

\begin{tabularx}{\linewidth}{l c c c c c c c}
\toprule
Method & LLM & Stage & \#Camera$\times$\#Time & $\mathrm{minADE}_6 \downarrow$  
& \#Scene tokens $\downarrow$ 
& \makecell{Training Time$\downarrow$\\(A100 GPU hrs)} 
& \makecell{Throughput$\uparrow$\\(clips/s)} \\
\midrule
$^*$\textit{Triplane (8-8-8)}~\cite{ivanovic2025efficient} & \textit{Llama2-1B} & \textit{Stage 1} & $4\times6$ & \textit{1.046} & \textit{1080 (1.2$\times$)} & \textit{4984 (2.2$\times$)} & \textit{69.62 (3.7$\times$)} \\
$^*$\textit{Triplane (4-6-6)}~\cite{ivanovic2025efficient} & \textit{Llama2-1B} & \textit{Stage 1} & $4\times6$ & \textit{0.974} & \textit{2496 (2.8$\times$)} & \textit{7960 (3.5$\times$)} & \textit{64.53 (3.5$\times$)} \\
In-house VLA & Qwen2-0.5B & Stage 1 & $2\times9$ & 0.818 & 2880 (3.2$\times$) & 4134 (1.8$\times$) & 18.60 (1.0$\times$) \\
\rowcolor{gray!20} \method (Ours) & Qwen2-0.5B & Stage 1 & $2\times9$ & \textbf{0.794} & \textbf{900} (1.0$\times$) & \textbf{2260} (1.0$\times$) & \textbf{41.08} (2.2$\times$) \\
\hline
In-house VLA & Qwen2-0.5B & Stage 1 + Stage 2 & $2\times9$ & 0.798 & 2880 (3.2$\times$) & 5750 (1.7$\times$) & 18.60 (1.0$\times$) \\
\rowcolor{gray!20} \method (Ours)  & Qwen2-0.5B & Stage 1 + Stage 2 & $2\times9$ & \textbf{0.761} & \textbf{900} (1.0$\times$) & \textbf{3318} (1.0$\times$) & \textbf{41.08} (2.2$\times$) \\
\bottomrule
\end{tabularx}

\vspace{0.1em}

$^*$We include results from \cite{ivanovic2025efficient} mainly for reference as there are many differences in experimental setup, including LLM training (from scratch in~\cite{ivanovic2025efficient} vs pretrained), non-interleaved~\cite{ivanovic2025efficient} vs interleaved prediction, and different choices of cameras $\times$ timesteps ($4 \times 6$ in~\cite{ivanovic2025efficient}).

\vspace{-2em}

\end{table*}

\subsection{Interleaved Prediction}
Next, we introduce an important design choice for VLA models in autonomous driving, named interleaved prediction.

\begin{figure}
    \centering
    \includegraphics[width=1\linewidth]{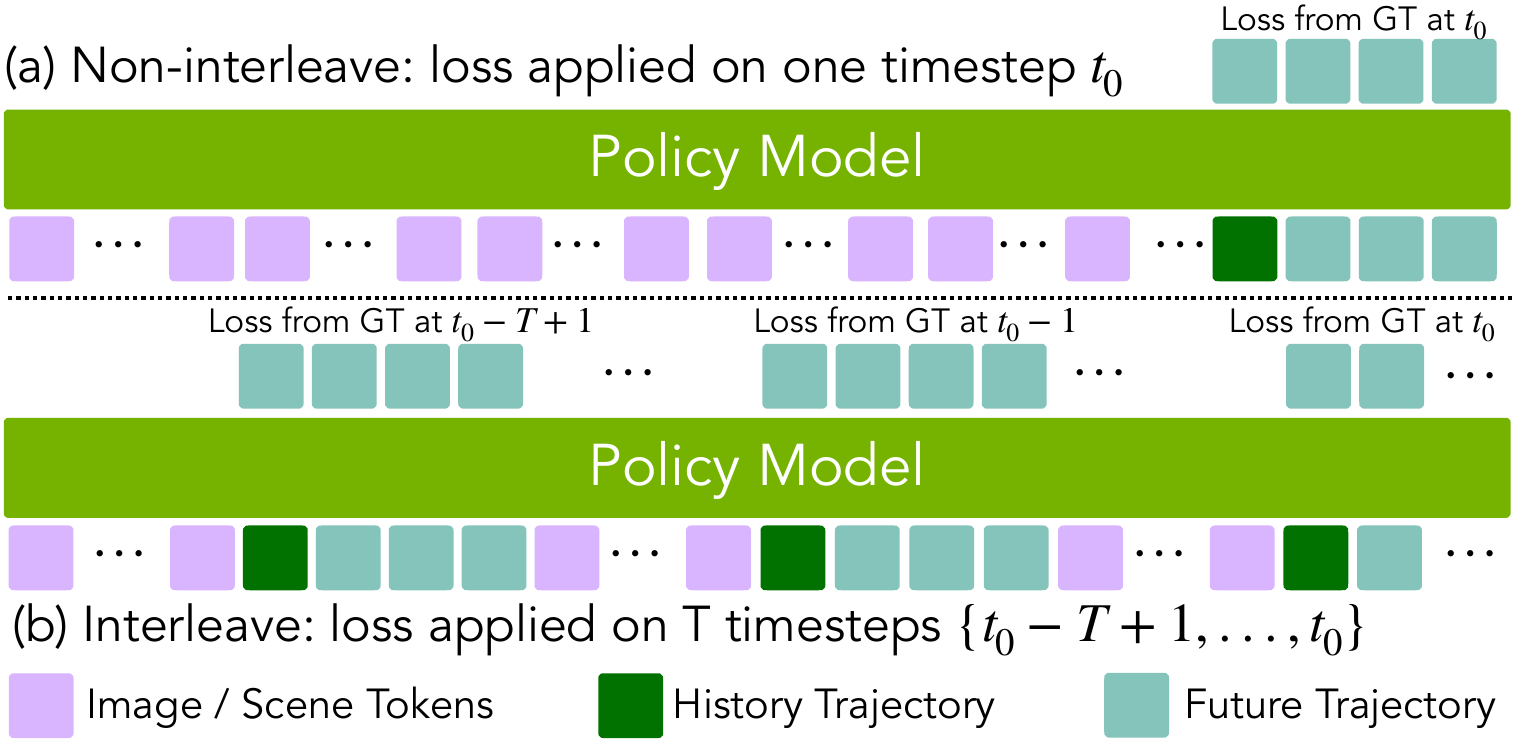}
    \vspace{-1.5em}
    \caption{Comparison of non-interleaved training strategies. (a) Loss applied only at the final timestep $t_0$. (b) Our proposed non-interleave setting: Loss applied across $T$ timesteps $\{t_0 - T + 1, \dots, t_0\}$. Special tokens are abbreviated for simplicity. GT denotes ground-truth.}
    \label{fig:interleave}
    \vspace{-1.2em}
\end{figure}

\paragraph{Diversifying Supervision.} At inference time, the policy model conditions on $C$ cameras over $T$ timesteps together with a single history token, and generates the predicted trajectory. A straightforward training setup would mirror this setting: the model takes the same number of timesteps context plus one history token and minimizes loss only on the subsequent future tokens, as shown in Fig.~\ref{fig:interleave}-(a). However, this turns out to be ineffective in practice (Tab.~\ref{tab:interleave_settings}). The model receives limited supervision.

We instead propose an interleaved training strategy, as shown in Fig.~\ref{fig:interleave}-(b). For a sequence of length $T$, we apply supervision on input contexts of varying lengths: at each timestep $k$, the model is given all \textit{image/scene} tokens up to $k$ and the corresponding trajectory history at $t_k$, and it is required to predict the next $H$ future steps. This produces $T$ supervision signals per sequence, one from every possible prefix. In practice, this can be implemented efficiently by adjusting the attention mask, without splitting the sequence. This interleaved design forces the model to predict the future with incomplete context, greatly increasing the diversity of training signals and leading to more efficient learning. We modify the attention mask so that the past history and future trajectory tokens in the context (e.g., from $t_{0}-T+1$ to $t_{0}-1$ in Fig.~\ref{fig:interleave}-b) do not contribute to the final prediction.

\paragraph{Adopting Holistic Scene Representations.}
Our \method representation $\mathbf{S}_{\text{\method}}$ is a single, jointly-encoded representation so that it no longer has an explicit mapping to a specific camera or time. To resolve this, we introduce a remarkably effective heuristic. We evenly partition the $K$ scene tokens into $T$ sequential, non-overlapping chunks, $\{\mathbf{S}^1_{\text{\method}}, \mathbf{S}^2_{\text{\method}}, \dots, \mathbf{S}^T_{\text{\method}}\}$, where each chunk $\mathbf{S}^k_{\text{\method}} \in \mathbb{R}^{(K/T) \times D_{llm}}$.

During interleaved training at step $k$, the policy model is conditioned on the concatenation of the first $k$ chunks of scene tokens, i.e., $\text{Concat}(\mathbf{S}^1_{\text{\method}}, \dots, \mathbf{S}^k_{\text{\method}})$, along with the most recent history token $\mathbf{h}_{hist}$. While this partitioning provides no explicit guidance to the encoder, the end-to-end training objective forces the model to learn a meaningful allocation. This process forces an emergent specialization within the scene tokens, where different chunks implicitly learn to capture information needed for the decision-making at every step, as shown in Fig.~\ref{fig:decomposition}.

\section{Experiments}

\subsection{Setup}

\paragraph{Dataset.}
We train and evaluate on a large-scale, internal dataset comprising \textbf{20,000 hours} of driving logs. The data was collected from a fleet of ego-vehicles operating in over 1,700 cities across 25 countries, ensuring high diversity. It encompasses a wide range of driving scenarios, including urban and highway driving, various weather patterns, different times of day, and diverse traffic densities. We use a geographically-separate holdout for evaluation, with a 90\% training/validation and 10\% test split. While ego-vehicles are equipped with seven cameras, we primarily utilize a two-camera setup (front-wide and front-telephoto) for efficiency. All videos are resized to a resolution of $320 \times 512$. We also present ablation studies on the number of camera inputs to demonstrate the wide applicability of our approach.

\paragraph{Training and Evaluation Details.} 
We adopt a two-stage training strategy. In the first stage, we freeze the image patchifier and train the scene encoder and policy model for 300k iterations (100k for ablation studies) using AdamW optimizer with a linear warmup over the first 1000 iterations to a peak of $4 \times 10^{-4}$, followed by a cosine decay schedule. We train with a global batch size of 256 (4 clips/GPU on 64 GPUs). The second stage consists of end-to-end fine-tuning, where all model parameters are unfrozen. For this stage, we reduce the learning rate to $1 \times 10^{-5}$ and train for 50k iterations. We sample observation windows of 2\,s during training and 1\,s during evaluation while keeping a fixed temporal sequence length of 9 timesteps. We train with standard mixed precision and enable activation checkpointing in the policy model. By default, we use a total of 900 scene tokens for 18 input images (2 cameras $\times$ 9 timesteps) and an 8-layer scene encoder with full self-attention.

\paragraph{Metrics.}
To provide a comprehensive comparison, we evaluate both the \textbf{efficiency} and \textbf{effectiveness} of our method against baselines. Efficiency is measured in terms of clips processed per second during inference and effectiveness is measured by driving performance using the standard $\text{minADE}_{k}$ metric, which is the minimum Average Displacement Error over $k$ predicted trajectories. We use $k=6$:
\begin{equation}
    \text{minADE}_{6}(Y, \hat{Y}) = \min_{k \in \{1,..,6\}} \frac{1}{H} \sum_{t=1}^{H} \left\| \hat{y}_t^{(k)} - y_t \right\|_2
\end{equation}
where $y_t$ is the ground truth ego-vehicle position at future timestep $t$, $\hat{y}_t^{(k)}$ is the $k$-th predicted position, and $H$ is the number of timesteps in the prediction horizon. 
We further average $\text{minADE}_{6}$ across time by 0.5s, 1.0s, 3.0s, and 5.0s.
% This metric evaluates the most optimistic prediction, rewarding multi-modal planners that can generate at least one accurate future path.

\paragraph{Baselines.} We compare our method against our in-house VLA model described in Section~\ref{sec:va_model}, which directly feeds the full, flattened sequence of all multi-view and multi-timestep image tokens into the LLM policy model. This comparison allows us to directly measure the efficiency gains and performance changes from introducing a compression bottleneck. Both our method and the baseline use the same policy head (Qwen2-0.5B~\cite{team2024qwen2}), trajectory tokenizers, and training schedule unless noted.

\subsection{Comparisons with Prior Systems}

We compare \method against prior state-of-the-art systems on our held-out test set. The results, presented in Table 1, demonstrate that our approach achieves a superior trade-off between efficiency and effectiveness.

Compared to our state-of-the-art in-house VLA model, \method achieves a 2.2$\times$ increase in inference throughput while simultaneously improving the driving performance ($\text{minADE}_6$) from 0.798 to 0.761. \method requires only about 60\% of the total training time compared to the baseline. This highlights the severe inefficiency of passing thousands of redundant tokens to the policy head.

\subsection{Ablation Study}
For ablation study, we train all models following stage-1 strategy and reduce the training iterations from 300k to 100k for manageable computation resources. By default, we use DINOv2-base as the image patchifier, 50 scene tokens per image (900 per scene), 8 scene encoder layers with joint self-attention, and 2 cameras (front-wide and front-telephoto). We analyze the effect of each component in the following sections. Our baseline follows the identical setting (same DINOv2-base backbone, LLM-policy head, and 100k training iterations).

\begin{figure*}[t]
  \centering

  % -------- Row 1: four plots --------
  \begin{subfigure}[b]{0.24\linewidth}
    \centering
    \includegraphics[width=\linewidth]{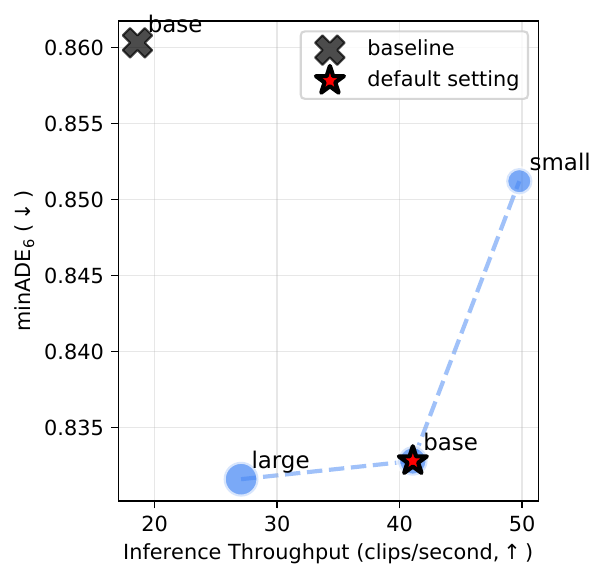}
    \caption{\textbf{Ablation on patchifier size}.}
    \label{fig:pachifier}
  \end{subfigure}\hfill
  \begin{subfigure}[b]{0.24\linewidth}
    \centering
    \includegraphics[width=\linewidth]{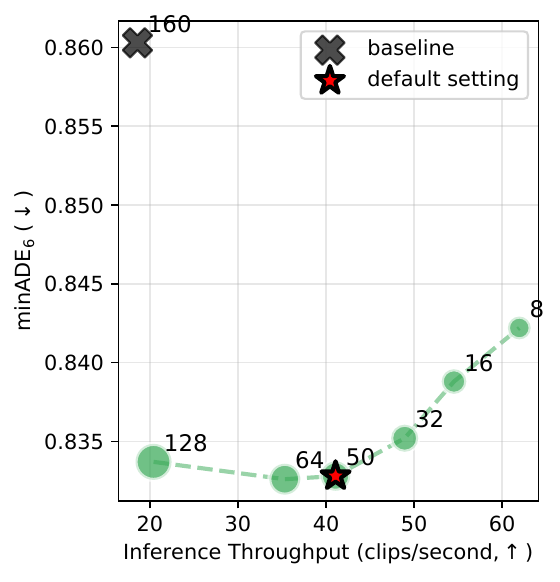}
    \caption{\textbf{Ablation on token number}.}
    \label{fig:token_count}
  \end{subfigure}\hfill
  \begin{subfigure}[b]{0.24\linewidth}
    \centering
    \includegraphics[width=\linewidth]{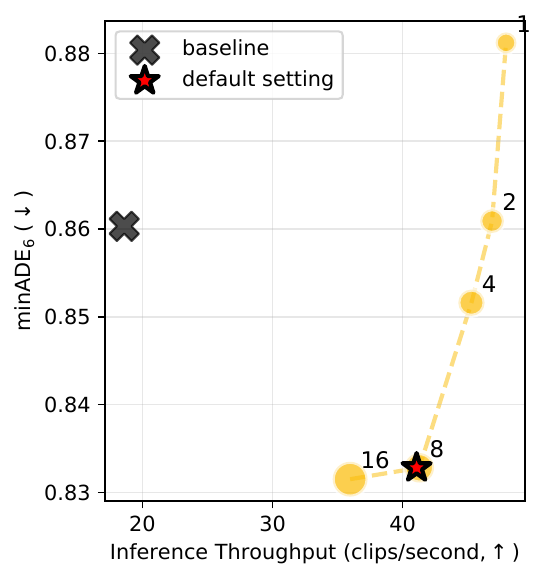}
    \caption{\textbf{Ablation on scene enc. layers}.}
    \label{fig:layers}
  \end{subfigure}\hfill
  \begin{subfigure}[b]{0.24\linewidth}
    \centering
    \includegraphics[width=\linewidth]{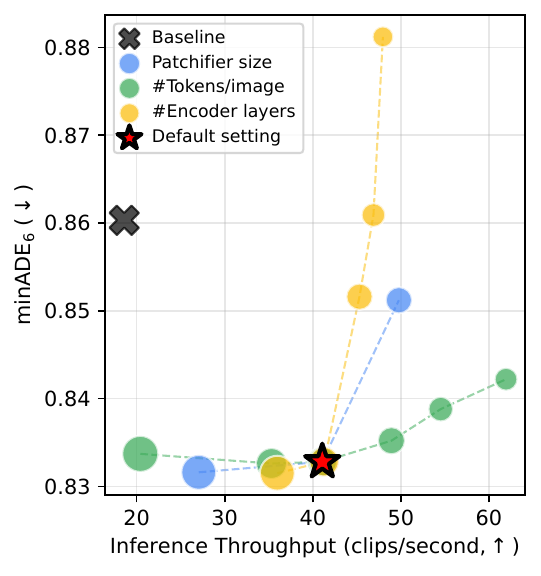}
    \caption{\textbf{Pareto frontier across settings}.}
    \label{fig:all_ablations}
  \end{subfigure}
  \vspace{-0.3em}

  % -------- Row 2: three tables --------
  \begin{subfigure}[b]{0.32\linewidth}
    \centering
    \caption{\textbf{Ablation on attention type}.}
    \label{tab:attention_type}
    % \vspace{0.25em}
    \small
    \resizebox{\linewidth}{!}{
    \begin{tabular}{lcc}
      \toprule
      Attention type & minADE$_6$ ($\downarrow$) & Throughput $\uparrow$ \\
      \midrule
      Per-Img. Cross-Attn   & 0.966    &  \textbf{47.96}  \\
      Per-Img. Joint-Attn   & 0.844     & 47.07   \\
      Cross-Attn  & 1.032  & 47.27 \\
      \rowcolor{gray!20} Joint-Attn & \textbf{0.833} & 41.08 \\
      \bottomrule
    \end{tabular}
    }
  \end{subfigure}\hfill
  \begin{subfigure}[b]{0.32\linewidth}
    \centering
    \caption{\textbf{Ablation on interleave settings}.}
    \label{tab:interleave_settings}
    % \vspace{0.25em}
    \small
    \resizebox{\linewidth}{!}{
    \begin{tabular}{lcc}
      \toprule
      setting & minADE$_6$ ($\downarrow$) & Throughput $\uparrow$ \\
      \midrule
      baseline, non-interleave & 0.894 & 18.47 \\
      baseline, interleave     & 0.860 & 18.60 \\
      Ours, non-interleave & 0.991 & 41.06 \\
      \rowcolor{gray!20} Ours, interleave     & \textbf{0.833} & \textbf{41.08} \\
      \bottomrule
    \end{tabular}
    }
  \end{subfigure}\hfill
  \begin{subfigure}[b]{0.32\linewidth}
    \centering
    \caption{\textbf{Ablation on camera count}.}
    \label{tab:cam_count}
    % \vspace{0.25em}
    \small
    \resizebox{\linewidth}{!}{
    \begin{tabular}{lcc}
      \toprule
      (camera $\times$ time)  & minADE$_6$ ($\downarrow$) & Throughput $\uparrow$ \\
      \midrule
      baseline (2$\times$9) & 0.860 & 18.60 \\
      baseline$^\dagger$ (4$\times$9) & 0.886 & 7.58 \\
      baseline$^\dagger$ (7$\times$9) & 0.925 &  3.34 \\
      \rowcolor{gray!20} Ours (2$\times$9), 50 tokens/img & 0.833 & \textbf{41.08} \\
      Ours (4$\times$9), 50 tokens/img & 0.831 & 19.06  \\
      Ours (7$\times$9), 32 tokens/img & \textbf{0.830} & 11.40 \\
      \bottomrule
    \end{tabular}
    }
  \end{subfigure}
  \vspace{-.5em}
  \caption{\textbf{Ablation study on \method{} design.}
    We evaluate key components of \method, including patchifier size, token count, scene encoder depth, attention type, interleave setting, and camera count. The default configuration (red star or gray row) achieves a strong accuracy–efficiency trade-off. Across all settings, \method consistently improves both accuracy and throughput (2$\times$–3$\times$) compared to the baseline. Each data point for our approach requires roughly 650 to 1,000 A100 GPU hours for training while baselines take around 1,300 to 1,800 GPU hours. $^\dagger$: Trained with half the batch size due to out-of-memory constraints.}
    \vspace{-1.3em}
    \label{fig:comprehensive_ablation}
\end{figure*}

\paragraph{Impact of patchifier size.} We evaluate the impact of the image patchifier by experimenting with three sizes of DINOv2~\cite{oquab2023dinov2}: Small, Base, and Large while keeping the scene encoder fixed. As shown in Fig.~\ref{fig:pachifier}, scaling from Small to Base yields a significant improvement in driving performance with a moderate increase in encoding latency. Moving from Base to Large yields more accuracy gains with more computational cost. We therefore adopt DINOv2-Base for all other experiments, as it offers the best balance of performance and efficiency.

\paragraph{Number of scene tokens.} 
We study how performance and efficiency scale with the number of scene tokens (K), varying it from 144 (8 tokens/image) to 1152 (128 tokens/image). Fig.~\ref{fig:token_count} shows a clear trade-off. Inference throughput is inversely proportional to K, as the cost of the policy model scales with the number of input tokens. Conversely, driving performance ($\mathrm{minADE}_6$) improves nearly monotonically as more scene tokens are used, saturating around $K\approx 900$. This yields a smooth Pareto frontier, enabling practitioners to select $K$ for a target throughput—an advantage over rigid image-based encoding. $K=900$ provides a good trade-off.

\paragraph{Number of scene encoder layers.} We ablate the depth of the \method scene encoder, testing configurations with 1, 2, 4, 8 and 16 Transformer layers. Fig.~\ref{fig:layers} shows the results. Performance improves notably when moving from 1 to 8 layers, suggesting that sufficient depth is needed for the scene tokens to effectively extract information from the thousands of image tokens, but excessively deep encoders provide little additional benefit. We therefore adopt 8 layers by default.

\paragraph{Pareto frontier across design choices.} Fig. \ref{fig:all_ablations} summarizes the ablation results by plotting the Pareto frontier across different design choices. Each curve traces the trade-off between accuracy and throughput when varying a single factor. The frontier clearly shows that our default configuration (red star) lies near the optimal trade-off region, offering strong driving performance without sacrificing efficiency.

\paragraph{Scene encoder design.} We compare four designs with the same output token budget K: 
\textbf{(1) Independent per-image tokens with cross- or self-attention}. Each camera image is independently compressed using its own set of learnable tokens before being passed to the policy model. This is efficient but ignores cross-spacetime redundancy. We evaluate both cross-attention layers, as used in Q-former~\cite{li2023blip}, and self-attention layers in this setting.
\textbf{(2) Joint scene tokens with cross- or self-attention}. A shared set of scene tokens is introduced to summarize information from all camera views and timesteps. In the cross-attention setting, scene tokens attend to the image tokens, but the image tokens themselves remain fixed (i.e., no self-attention among image tokens or between image and scene tokens). We study both cross- and self-attention mechanisms to understand the importance of updating scene tokens and image tokens.
Table~\ref{tab:attention_type} summarizes the results. First, the widely employed per-image token strategy performs worse than joint encoding. This is expected, as it cannot leverage mutual information exist across different camera views and timesteps to obtain a holistic scene representation. Second, within both families, the cross-attention variants underperform their self-attention counterparts. These results suggest that enabling image tokens to interact with one another, mediated through scene tokens, is critical for learning effective representations. Overall, a \textit{holistic joint encoding} method proves essential to the success of \method.

\paragraph{Interleaved prediction.} Table~\ref{tab:interleave_settings} shows that interleaving is crucial for both baseline and \method. For the baseline model, interleaving reduces $\text{minADE}_6$ from 0.894 to 0.860, with no loss in efficiency. For \method, the effect is even stronger: interleaving improves $\text{minADE}_6$ from 0.991 to 0.833 while preserving high throughput (41 clips/s). This training strategy forces the model to predict trajectories from partial contexts rather than always conditioning on the full history. It also provides richer supervision and greater robustness because of the usage of intermediate ground truth. Overall, interleaving is key to enabling sample-efficient learning and unlocking the full benefits of our scene-token encoder.

\begin{figure*}[h!]
    \centering
    \includegraphics[width=1\linewidth]{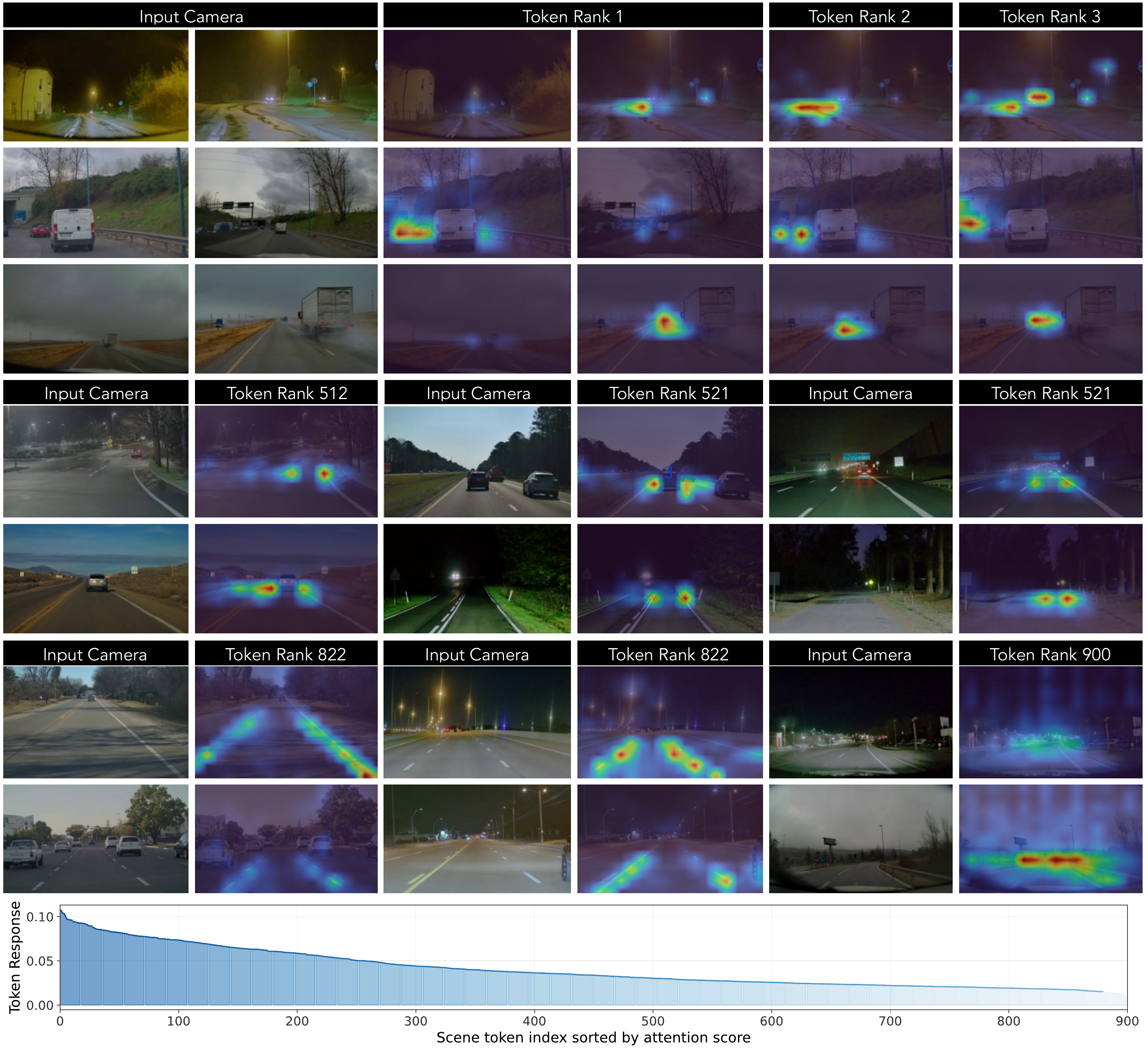}
  \vspace{-1.7em}
    \caption{\textbf{Emergent scene decomposition from learned scene tokens.} \textbf{Bottom}: We show the sorted average token response for all scene tokens. Larger value indicates that these tokens have stronger attention to image tokens. \textbf{Top.} Visualization of representative scene tokens. \textit{Top three rows}: the highest-ranked tokens (Ranks 1–3) consistently focus on the ego vehicle’s destination. \textit{Middle two rows}: mid-ranked tokens (e.g., Rank 512) exhibit interesting look-ahead scanning patterns across the roadway, resembling human's visual scanning, guiding fixation or look-ahead fixation strategies. \textit{Next two rows}: a lower-ranked token (Rank 822) clearly highlights road lane markings. Interestingly, the least responsive token (Rank 900) appears to capture positional bias patterns. Notably, all these behaviors emerge naturally without explicit supervision.}
    \label{fig:decomposition}
  \vspace{-1.2em}
\end{figure*}

\paragraph{The sensitivity to camera count.} 
We further examine the impact of increasing the number of input cameras (Table~\ref{tab:cam_count}). Our method continues to provide consistent gains over baselines in both efficiency and effectiveness. In particular, the throughput advantage grows from 2.2$\times$ with 2 cameras to 3.4$\times$ with 7 cameras, while maintaining strong driving performance. Notably, as the number of cameras increases, baseline performance degrades significantly, whereas our method remains stable. We attribute this to two factors. First, learning effective policies from tens of thousands of raw image tokens is increasingly challenging. Second, the front-wide and front-telephoto cameras already provide sufficient coverage for most driving scenarios in our test split, which does not specifically emphasize rare cases requiring side or rear cameras. Exploring these scenarios remains an interesting direction for future work.

\subsection{Emerging Scene Decomposition and Intention Tokens}

To probe what the scene tokens capture, we analyze their attention toward image tokens by inspecting the attention weights produced by the scene encoder. Specifically, we extract the attention from scene tokens to image tokens at the final layer of the scene encoder, average these values across attention heads, and then record the maximal response from each scene token to any image token. This procedure yields a per-scene-token measure of its strongest attentional link, which we use for quantitative analysis and visualization of representative examples in Fig.~\ref{fig:decomposition}.

The results show clear specialization. The top-ranked tokens (Rank 1-3) consistently highlight the destination of the ego vehicle, suggesting that the model dedicates capacity to goal-related reasoning. Mid-ranked tokens (Rank 521) exhibit scanning behaviors (two-spot attention), shifting their attention forward along the road in a manner resembling look-ahead planning. Lower-ranked tokens (Rank 822) capture static yet crucial structures such as lane markings. We also note that there are more tokens related to lane markings, meaning that these scene tokens work collectively. At the tail end, the least responsive token (Rank 900) appears to encode positional or bias-related patterns. Its semantics are unclear, but similar phenomena have been observed in ViT~\cite{dosovitskiy2020image}, where certain tokens serve structural roles (e.g., encoding positions) rather than explicit content~\cite{yang2023emernerf,yang2024denoising,darcet2023vision}.

Overall, this analysis demonstrates that the scene encoder does more than compress visual inputs: it learns to allocate tokens to semantically meaningful components, including destination, motion anticipation, road layout, and positional context. The emergence of intention-like tokens, often with \textit{benign} redundancy in critical regions such as the destination and landmarks, while devoting fewer resources to less informative areas like the sky or surrounding trees. Importantly, these behaviors arise end-to-end without handcrafted priors or explicit supervision, showing that joint attention between scene and image tokens is sufficient to induce meaningful specialization. This emergent decomposition partially demonstrates why the representation remains compact yet effective, enabling \method to achieve high performance with significantly fewer tokens.

\section{Conclusion}

In this work, we introduced \method, a simple, efficient, and data-driven scene encoder for end-to-end autonomous driving. Our core contribution is a lightweight, geometry-agnostic Transformer that jointly encodes multi-view, multi-timestep images into a compact set of learnable scene tokens. Our extensive studies confirm that the joint, holistic nature of our encoder is critical to its success and that it is flexible with respect to model size and sensor configuration and that our compressed scene tokens learn an emergent, semantically meaningful decomposition of the driving scene. Exploring the interpretability and the specialization of these implicit scene tokens would be an intriguing future direction. \textbf{Limitations.}
While \method achieves strong efficiency–accuracy trade-offs, we note that it inherits the biases and coverage limitations of the proprietary driving dataset, which may underrepresent rare long-tail scenarios. Broader evaluation on more diverse and careful consideration of downstream social impact, including safety and fairness across diverse geographies, remain important directions for future work.

\vspace{-.5em}
\bibliographystyle{IEEEtran}
\bibliography{icra26}

% Generated by IEEEtran.bst, version: 1.14 (2015/08/26)
\begin{thebibliography}{10}
\providecommand{\url}[1]{#1}
\csname url@samestyle\endcsname
\providecommand{\newblock}{\relax}
\providecommand{\bibinfo}[2]{#2}
\providecommand{\BIBentrySTDinterwordspacing}{\spaceskip=0pt\relax}
\providecommand{\BIBentryALTinterwordstretchfactor}{4}
\providecommand{\BIBentryALTinterwordspacing}{\spaceskip=\fontdimen2\font plus
\BIBentryALTinterwordstretchfactor\fontdimen3\font minus \fontdimen4\font\relax}
\providecommand{\BIBforeignlanguage}[2]{{%
\expandafter\ifx\csname l@#1\endcsname\relax
\typeout{** WARNING: IEEEtran.bst: No hyphenation pattern has been}%
\typeout{** loaded for the language `#1'. Using the pattern for}%
\typeout{** the default language instead.}%
\else
\language=\csname l@#1\endcsname
\fi
#2}}
\providecommand{\BIBdecl}{\relax}
\BIBdecl

\bibitem{brohan2022rt}
A.~Brohan, N.~Brown, J.~Carbajal, Y.~Chebotar, J.~Dabis, C.~Finn, K.~Gopalakrishnan, K.~Hausman, A.~Herzog, J.~Hsu, J.~Ibarz, B.~Ichter, A.~Irpan, T.~Jackson, S.~Jesmonth, N.~J. Joshi, R.~Julian, D.~Kalashnikov, Y.~Kuang, I.~Leal, K.-H. Lee, S.~Levine, Y.~Lu, U.~Malla, D.~Manjunath, I.~Mordatch, O.~Nachum, C.~Parada, J.~Peralta, E.~Perez, K.~Pertsch, J.~Quiambao, K.~Rao, M.~S. Ryoo, G.~Salazar, P.~R. Sanketi, K.~Sayed, J.~Singh, S.~Sontakke, A.~Stone, C.~Tan, H.~Tran, V.~Vanhoucke, S.~Vega, Q.~Vuong, F.~Xia, T.~Xiao, P.~Xu, S.~Xu, T.~Yu, and B.~Zitkovich, ``Rt-1: Robotics transformer for real-world control at scale,'' in \emph{RSS}, 2023.

\bibitem{kim2024openvla}
M.~J. Kim, K.~Pertsch, S.~Karamcheti, T.~Xiao, A.~Balakrishna, S.~Nair, R.~Rafailov, E.~P. Foster, P.~R. Sanketi, Q.~Vuong, T.~Kollar, B.~Burchfiel, R.~Tedrake, D.~Sadigh, S.~Levine, P.~Liang, and C.~Finn, ``Openvla: An open-source vision-language-action model,'' in \emph{CoRL}, 2024.

\bibitem{gao2024survey}
H.~Gao, Z.~Wang, Y.~Li, K.~Long, M.~Yang, and Y.~Shen, ``A survey for foundation models in autonomous driving,'' \emph{arXiv preprint arXiv:2402.01105}, 2024.

\bibitem{tian2025drivevlm}
X.~Tian, J.~Gu, B.~Li, Y.~Liu, Y.~Wang, Z.~Zhao, K.~Zhan, P.~Jia, X.~Lang, and H.~Zhao, ``Drivevlm: The convergence of autonomous driving and large vision-language models,'' in \emph{CoRL}, 2024.

\bibitem{ivanovic2025efficient}
B.~Ivanovic, C.~Saltori, Y.~You, Y.~Wang, W.~Luo, and M.~Pavone, ``Efficient multi-camera tokenization with triplanes for end-to-end driving,'' \emph{IEEE Robotics and Automation Letters}, vol.~10, no.~11, pp. 11\,713--11\,720, 2025.

\bibitem{li2024bevformer}
Z.~Li, W.~Wang, H.~Li, E.~Xie, C.~Sima, T.~Lu, Q.~Yu, and J.~Dai, ``Bevformer: learning bird's-eye-view representation from lidar-camera via spatiotemporal transformers,'' \emph{IEEE Transactions on Pattern Analysis and Machine Intelligence}, 2024.

\bibitem{hu2023planning}
Y.~Hu, J.~Yang, L.~Chen, K.~Li, C.~Sima, X.~Zhu, S.~Chai, S.~Du, T.~Lin, W.~Wang \emph{et~al.}, ``Planning-oriented autonomous driving,'' in \emph{CVPR}, 2023.

\bibitem{tong2023scene}
W.~Tong, C.~Sima, T.~Wang, L.~Chen, S.~Wu, H.~Deng, Y.~Gu, L.~Lu, P.~Luo, D.~Lin \emph{et~al.}, ``Scene as occupancy,'' in \emph{ICCV}, 2023.

\bibitem{huang2023tri}
Y.~Huang, W.~Zheng, Y.~Zhang, J.~Zhou, and J.~Lu, ``Tri-perspective view for vision-based 3d semantic occupancy prediction,'' in \emph{CVPR}, 2023.

\bibitem{cao2023hexplane}
A.~Cao and J.~Johnson, ``Hexplane: A fast representation for dynamic scenes,'' in \emph{CVPR}, 2023.

\bibitem{darcet2023vision}
T.~Darcet, M.~Oquab, J.~Mairal, and P.~Bojanowski, ``Vision transformers need registers,'' in \emph{ICLR}, 2024.

\bibitem{vaswani2017attention}
A.~Vaswani, N.~Shazeer, N.~Parmar, J.~Uszkoreit, L.~Jones, A.~N. Gomez, {\L}.~Kaiser, and I.~Polosukhin, ``Attention is all you need,'' \emph{NeurIPS}, 2017.

\bibitem{li2023blip}
J.~Li, D.~Li, S.~Savarese, and S.~Hoi, ``Blip-2: Bootstrapping language-image pre-training with frozen image encoders and large language models,'' in \emph{ICML}, 2023.

\bibitem{hu2022st}
S.~Hu, L.~Chen, P.~Wu, H.~Li, J.~Yan, and D.~Tao, ``St-p3: End-to-end vision-based autonomous driving via spatial-temporal feature learning,'' in \emph{ECCV}, 2022.

\bibitem{chen2024vadv2}
S.~Chen, B.~Jiang, H.~Gao, B.~Liao, Q.~Xu, Q.~Zhang, C.~Huang, W.~Liu, and X.~Wang, ``Vadv2: End-to-end vectorized autonomous driving via probabilistic planning,'' \emph{arXiv preprint arXiv:2402.13243}, 2024.

\bibitem{liu2022petr}
Y.~Liu, T.~Wang, X.~Zhang, and J.~Sun, ``Petr: Position embedding transformation for multi-view 3d object detection,'' in \emph{ECCV}, 2022.

\bibitem{liu2023petrv2}
Y.~Liu, J.~Yan, F.~Jia, S.~Li, A.~Gao, T.~Wang, and X.~Zhang, ``Petrv2: A unified framework for 3d perception from multi-camera images,'' in \emph{ICCV}, 2023.

\bibitem{liang2022bevfusion}
T.~Liang, H.~Xie, K.~Yu, Z.~Xia, Z.~Lin, Y.~Wang, T.~Tang, B.~Wang, and Z.~Tang, ``Bevfusion: A simple and robust lidar-camera fusion framework,'' in \emph{NeurIPS}, 2022.

\bibitem{wei2023surroundocc}
Y.~Wei, L.~Zhao, W.~Zheng, Z.~Zhu, J.~Zhou, and J.~Lu, ``Surroundocc: Multi-camera 3d occupancy prediction for autonomous driving,'' in \emph{ICCV}, 2023.

\bibitem{chan2022efficient}
E.~R. Chan, C.~Z. Lin, M.~A. Chan, K.~Nagano, B.~Pan, S.~De~Mello, O.~Gallo, L.~J. Guibas, J.~Tremblay, S.~Khamis \emph{et~al.}, ``Efficient geometry-aware 3d generative adversarial networks,'' in \emph{CVPR}, 2022.

\bibitem{fridovich2023k}
S.~Fridovich-Keil, G.~Meanti, F.~R. Warburg, B.~Recht, and A.~Kanazawa, ``K-planes: Explicit radiance fields in space, time, and appearance,'' in \emph{CVPR}, 2023.

\bibitem{muller2022instant}
T.~M{\"u}ller, A.~Evans, C.~Schied, and A.~Keller, ``Instant neural graphics primitives with a multiresolution hash encoding,'' \emph{ACM transactions on graphics (TOG)}, 2022.

\bibitem{yang2023emernerf}
J.~Yang, B.~Ivanovic, O.~Litany, X.~Weng, S.~W. Kim, B.~Li, T.~Che, D.~Xu, S.~Fidler, M.~Pavone, and Y.~Wang, ``Emernerf: Emergent spatial-temporal scene decomposition via self-supervision,'' in \emph{ICLR}, 2024.

\bibitem{jaegle2021perceiver}
A.~Jaegle, F.~Gimeno, A.~Brock, O.~Vinyals, A.~Zisserman, and J.~Carreira, ``Perceiver: General perception with iterative attention,'' in \emph{ICML}, 2021.

\bibitem{jaegle2021perceiverio}
A.~Jaegle, S.~Borgeaud, J.-B. Alayrac, C.~Doersch, C.~Ionescu, D.~Ding, S.~Koppula, D.~Zoran, A.~Brock, E.~Shelhamer, O.~Hénaff, M.~M. Botvinick, A.~Zisserman, O.~Vinyals, and J.~Carreira, ``Perceiver io: A general architecture for structured inputs \& outputs,'' in \emph{ICLR}, 2022.

\bibitem{alayrac2022flamingo}
J.-B. Alayrac, J.~Donahue, P.~Luc, A.~Miech, I.~Barr, Y.~Hasson, K.~Lenc, A.~Mensch, K.~Millican, M.~Reynolds \emph{et~al.}, ``Flamingo: a visual language model for few-shot learning,'' in \emph{NeurIPS}, 2022.

\bibitem{ryoo2021tokenlearner}
M.~S. Ryoo, A.~Piergiovanni, A.~Arnab, M.~Dehghani, and A.~Angelova, ``Tokenlearner: What can 8 learned tokens do for images and videos?'' in \emph{NeurIPS}, 2021.

\bibitem{yu2024image}
Q.~Yu, M.~Weber, X.~Deng, X.~Shen, D.~Cremers, and L.-C. Chen, ``An image is worth 32 tokens for reconstruction and generation,'' in \emph{NeurIPS}, 2024.

\bibitem{jin2024lvsm}
H.~Jin, H.~Jiang, H.~Tan, K.~Zhang, S.~Bi, T.~Zhang, F.~Luan, N.~Snavely, and Z.~Xu, ``Lvsm: A large view synthesis model with minimal 3d inductive bias,'' in \emph{ICLR}, 2025.

\bibitem{yang2024storm}
J.~Yang, J.~Huang, Y.~Chen, Y.~Wang, B.~Li, Y.~You, M.~Igl, A.~Sharma, P.~Karkus, D.~Xu, B.~Ivanovic, Y.~Wang, and M.~Pavone, ``Storm: Spatio-temporal reconstruction model for large-scale outdoor scenes,'' in \emph{ICLR}, 2025.

\bibitem{oquab2023dinov2}
M.~Oquab, T.~Darcet, T.~Moutakanni, H.~V. Vo, M.~Szafraniec, V.~Khalidov, P.~Fernandez, D.~Haziza, F.~Massa, A.~El-Nouby \emph{et~al.}, ``Dinov2: Learning robust visual features without supervision,'' \emph{Transactions on Machine Learning Research}, 2023.

\bibitem{wang2025vggt}
J.~Wang, M.~Chen, N.~Karaev, A.~Vedaldi, C.~Rupprecht, and D.~Novotny, ``Vggt: Visual geometry grounded transformer,'' in \emph{CVPR}, 2025.

\bibitem{peebles2023scalable}
W.~Peebles and S.~Xie, ``Scalable diffusion models with transformers,'' in \emph{ICCV}, 2023.

\bibitem{team2024qwen2}
A.~Yang, B.~Yang, B.~Hui, B.~Zheng, B.~Yu, C.~Zhou, C.~Li, C.~Li, D.~Liu, F.~Huang, G.~Dong, H.~Wei, H.~Lin, J.~Tang, J.~Wang, J.~Yang, J.~Tu, J.~Zhang, J.~Ma, J.~Xu, J.~Zhou, J.~Bai, J.~He, J.~Lin, K.~Dang, K.~Lu, K.~Chen, K.~Yang, M.~Li, M.~Xue, N.~Ni, P.~Zhang, P.~Wang, R.~Peng, R.~Men, R.~Gao, R.~Lin, S.~Wang, S.~Bai, S.~Tan, T.~Zhu, T.~Li, T.~Liu, W.~Ge, X.~Deng, X.~Zhou, X.~Ren, X.~Zhang, X.~Wei, X.~Ren, Y.~Fan, Y.~Yao, Y.~Zhang, Y.~Wan, Y.~Chu, Y.~Liu, Z.~Cui, Z.~Zhang, Z.~Guo, and Z.~Fan, ``Qwen2 technical report,'' \emph{arXiv preprint arXiv:2407.10671}, 2024.

\bibitem{dosovitskiy2020image}
A.~Dosovitskiy, L.~Beyer, A.~Kolesnikov, D.~Weissenborn, X.~Zhai, T.~Unterthiner, M.~Dehghani, M.~Minderer, G.~Heigold, S.~Gelly, J.~Uszkoreit, and N.~Houlsby, ``An image is worth 16×16 words: Transformers for image recognition at scale,'' in \emph{ICLR}, 2021.

\bibitem{yang2024denoising}
J.~Yang, K.~Z. Luo, J.~Li, C.~Deng, L.~Guibas, D.~Krishnan, K.~Q. Weinberger, Y.~Tian, and Y.~Wang, ``Denoising vision transformers,'' in \emph{ECCV}, 2024.

\end{thebibliography}

\end{document}